\documentclass[11pt]{article}

\usepackage[final]{acl}

\usepackage{times}
\usepackage{latexsym}

\usepackage[T1]{fontenc}

\usepackage[utf8]{inputenc}

\usepackage{microtype}

\usepackage{inconsolata}

\usepackage{graphicx}
\usepackage{subcaption}

\usepackage{amsmath}
\usepackage{bbold}

\usepackage{multirow}

\usepackage{xcolor}

\title{Localizing Cross-Lingual Generalization in Parameter Space}

\author{
Yeoreum Lee\textsuperscript{1}, Jaeseong Lee\textsuperscript{2}, Seung-won Hwang\textsuperscript{1}\thanks{\ Corresponding author.} \\
Interdisciplinary Program in Artificial Intelligence, Seoul National University\textsuperscript{1} \\
Snowflake AI Research\textsuperscript{2}, \\
\texttt{leeyeoreum01@snu.ac.kr, jaeseong.lee@snowflake.com, seungwonh@snu.ac.kr} \\
}

\begin{document}
\maketitle
\begin{abstract}
Cross-lingual transfer is uneven across tasks, succeeding for natural language understanding but often failing for mathematical reasoning.
We argue that this asymmetry is governed by the geometry of parameter space.
Specifically, cross-lingual generalization is determined by whether English and target-language LoRA adapters share a low-error basin.
To test this, we use linear mode connectivity (LMC) between English and target-language adapters, evaluating on three tasks.
We find that NLI and QA model-language pairs are mostly linearly connected, whereas mathematical reasoning (Math) pairs are not, indicating that basin geometry is task-dependent.
Such disconnection in Math is ``local'' rather than global. A small target-language task sample, applied only to the English adapter's language encoding or semantic composition layers, is enough to repair it.
Cross-lingual generalization is also factorizable into language-specific encoding and language-agnostic composition, each supplied separately from a cheaper signal.
English task data routed to the composition layers and an unlabeled target-language corpus routed to the language encoding layers each alleviate it on their own for nearly every pair.
Composition is the more influential of these, and it needs no target-language data.
Together, these results suggest that cross-lingual generalization decomposes into language-specific and language-agnostic components in parameter space.
This decomposition offers a practical route for low-resource cross-lingual adaptation.
\end{abstract}

\section{Introduction}
Large language models (LLMs) have demonstrated remarkable capabilities across a wide range of NLP tasks, including natural language understanding and reasoning tasks~\citep{openai2023gpt4technicalreport,grattafiori2024llama3,gemmateam2024gemma2,yang2025qwen3,geminiteam2023geminifamilyhighlycapable}.
However, these capabilities remain largely English-centric.
This has driven growing interest in cross-lingual transfer, which leverages the strong capabilities of high-resource languages (e.g., English) to improve performance in other languages~\citep{lauscher2020zero,conneau2020unsupervised,hu2025large,bandarkar2025layer}.
Among these, a growing line of work performs cross-lingual transfer with LoRA for efficient adaptation to target languages~\citep{whitehouse2024low,zhao2025adamergex}.

But why is such cross-lingual transfer successful in some cross-lingual settings but not in others?
Prior work approaches this question either externally, through properties of the language pair, or internally, through the mechanisms of an isolated pre-trained model.
Specifically, the former predicts how well generalization will transfer between languages from properties of the language pair such as typological distance~\citep{lin2019choosing,philippy2023towards}.
However, since these properties are tied to language pairs rather than models, they cannot account for variation across different models or tasks for the same pair.
Alternatively, the latter probes the phenomenon internally, through neuron-level analyses~\citep{tang2024language,kojima2024multilingual} and representation-level analyses that characterize how representations evolve along the depth of the model~\citep{wendler2024llamas,wu2025the,zhong2025language}.
Yet these analyses only examine the internal mechanisms of an isolated pre-trained model and do not explicitly compare it against an expert model specialized for the target language.
We instead study cross-lingual generalization in parameter space, comparing the English adapter with a target-language expert.
This shifts the question from external linguistic properties or internal pre-trained representations to the geometry of adaptation itself.

A natural lens for this geometry is \emph{linear mode connectivity} (LMC), which evaluates whether two models fine-tuned from the same initialization share the same error basin~\citep{frankle2020linear,entezari2022the}.
Applied to English and target-language adapters, LMC indicates whether they lie within a connected low-error region, providing a parameter-space probe for cross-lingual generalization.
LMC has been studied for pre-trained language models~\citep{qin2022exploring} and multi-task LoRA adapters~\citep{zhuang2025come}, but not yet for the cross-lingual case.

We cover cross-lingual generalization across three models and three tasks spanning natural language understanding (NLI and QA) and mathematical reasoning (Math).
Our analysis surfaces three findings. First, basin geometry is task-dependent: NLI and QA model-language pairs are almost always linearly connected, while Math pairs largely fail to connect.
Second, this deficit is local rather than global: adapting either the early or the middle layers alone restores connectivity in Math, while the final layers contribute little.
Relative to the target-language adapter, this indicates that the English adapter falls short in input encoding and semantic composition, not in output decoding.
Third, cross-lingual generalization is factorizable into language-specific input encoding and language-agnostic semantic composition.
Each improves LMC on its own for nearly every pair, composition from English task data at the middle layers and language-specific processing from an unlabeled corpus at the early layers.
Neither signal requires target-language task labels.
Of the two, semantic composition is the more influential factor, and the language-specific signal adds little beyond it.
Together, these findings reveal cross-lingual generalization as the parameter-space alignment of two components that can be addressed separately through layer-targeted training.
In practice, English task supervision at the composition layers alone can extend an English adapter to a new target language, without any target-language data.

Our contributions are as follows:
\begin{itemize}
    \item A parameter-space characterization of cross-lingual generalization through LMC, applied across three LLMs, three tasks, and four target languages.
    \item Evidence that cross-lingual generalization is task-dependent rather than monolithic, with NLU adapter pairs connecting cleanly while Math pairs largely fail to connect.
    \item A layer-level localization of the generalization deficit, showing that adapting only the adapter's early or middle layers suffices to repair it.
    \item A parameter-space factorization of cross-lingual generalization into language-specific processing and language-agnostic composition.
    It is exploitable through layer-targeted training on English supervision alone, without any target-language data.
\end{itemize}

\section{Related work}
\paragraph{Cross-lingual transfer.}
A broad line of work has pursued cross-lingual transfer, improving its performance through various adaptation and merging strategies.
Early work demonstrated that multilingual language models achieve zero-shot cross-lingual transfer~\citep{pires2019multilingual,conneau2020unsupervised}.
Subsequent work has improved transfer performance through representation alignment~\citep{she2024mapo}, merging-based approaches that compose language and task modules in parameter space~\citep{ansell2022composable,zhao2025adamergex,bandarkar2025layer,chen2025extracting}, and leveraging internal English knowledge~\citep{etxaniz2024multilingual,zhu2024question}.
However, these studies build transfer by composing language and task modules.
We instead use parameter-space geometry to characterize when cross-lingual generalization holds, rather than to produce it.

\paragraph{Localizing cross-lingual generalization.}
What drives cross-lingual generalization has been analyzed through external linguistic features~\citep{lin2019choosing,ahuja2022multi,philippy2023towards,blaschke2025analyzing}.
\citet{riemenschneider2025cross} go further and argue that cross-lingual transfer is better treated as an evaluation method than as a capability in itself.
These studies describe transfer from the outside and leave open where, if anywhere, it is realized inside the model.
This motivates a line of work that asks where inside a pre-trained LLM cross-lingual generalization is realized, locating it along the depth of the network.

Neuron-level analyses suggest that this generalization is realized in the interior of the network.
Activations sensitive to a specific language tend to concentrate near the outer portions rather than the interior~\citep{tang2024language,kojima2024multilingual,zhu2024multilingual}.
Logit-lens and activation-patching studies localize generalization to the same interior region from a representational standpoint~\citep{wendler2024llamas,zhao2024how,schut2025do,wu2025the}.
They show that the inner forward pass operates in a language-agnostic (or English-aligned) space, with the target language emerging only near the output.
Some work argues that language-agnostic and language-specific representations can coexist along the same layers~\citep{chang2022geometry,brinkmann2025large,dumas2025separating,fierro2025multilingual}.
Other work finds that the dominant internal latent language is itself shaped by the pre-training mixture rather than defaulting to English~\citep{zhong2025language}.
However, all of these analyses operate within the internal representations and computations of the model.
They leave unexamined the parameter-space geometry between an isolated pre-trained model and an expert model specialized for the target language.
\vspace{-5pt}

\paragraph{Linear mode connectivity.}
A series of studies has established that models sharing the same initialization occupy the same error basin, giving rise to LMC as a tool for probing the structure of error landscapes~\citep{frankle2020linear,neyshabur2020what,entezari2022the}.
\citet{qin2022exploring} provided the first systematic LMC analysis of pre-trained language models.
Recent work has examined LMC of LoRA adapters in multi-task scenarios~\citep{zhuang2025come}.
However, all previous LMC studies in NLP compare endpoints trained on the same language data, leaving the geometry between English and target-language adapters outside their scope.

\section{Evaluation design for research questions}
\label{sec:setup}

\begin{figure*}[t]
  \centering
  \includegraphics[width=\linewidth]{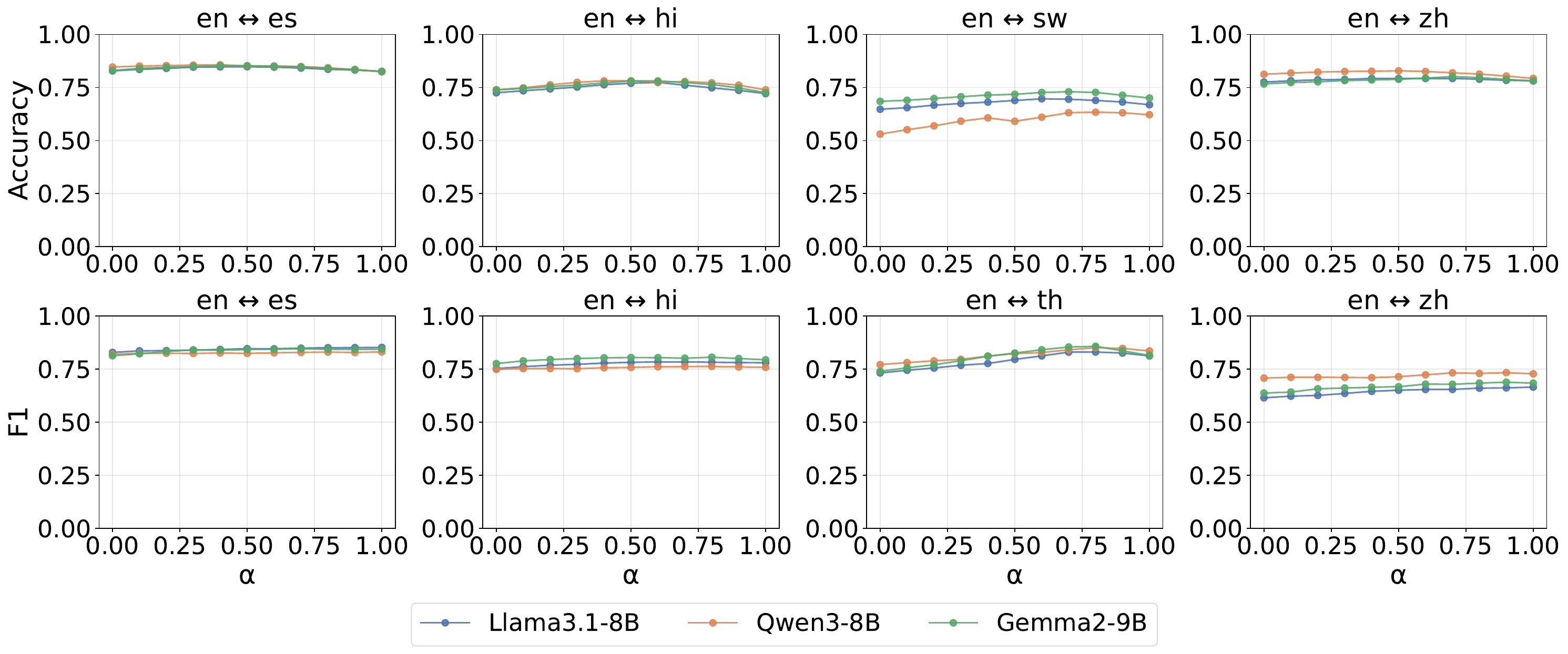}
  \caption{
    Linear interpolation paths between the English adapter $(\alpha = 0)$ and the target-language adapter $(\alpha = 1)$ for NLI and QA.
    The first row reports accuracy evaluated on XNLI, and the second row reports F1 evaluated on XQuAD. 
    Each subplot reports task performance in the respective target language along the interpolation path.
  }
  \label{fig:lmc_main}
  \vspace{-3pt}
\end{figure*}

\paragraph{Research questions.}
We investigate cross-lingual generalization through three questions, each established in a later section.
First, is cross-lingual generalization task-dependent rather than uniform across tasks (\S\ref{sec:R1})?
Second, when it breaks on Math, is the deficit layer-localized rather than global (\S\ref{sec:R2})?
Third, does the deficit factorize into two separately suppliable components (\S\ref{sec:R3})?
All three questions rest on the same parameter-space probe, which we lay out in the remainder of this section.

\paragraph{Tasks and datasets.}
Our experiments cover three tasks that range from NLU to mathematical reasoning: NLI, QA, and Math, with English as the pivot language.
Across all three tasks, we use datasets whose multilingual training and test sets are translations of a common English source.
This ensures that differences in basin structure across adapters reflect language rather than data content.

\paragraph{Models and training.}
To ensure that our analysis is not tied to a single model, we use three instruction-tuned LLMs of similar scale: Llama 3.1-8B-Instruct~\citep{grattafiori2024llama3}, Qwen3-8B~\citep{yang2025qwen3}, and Gemma 2-9B-it~\citep{gemmateam2024gemma2} as base models.
Each model is fine-tuned independently on English or corresponding target language using LoRA adapters~\citep{hu2022lora}, sampling 5,000 examples per language.
Target-language data is scarce in practice, so an adapter fine-tuned directly on such data is the ideal comparison endpoint for the English adapter.
All adapters are trained for 3 epochs using LlamaFactory~\citep{zheng2024llamafactory}.
Further experimental details are provided in Appendix~\ref{appdix_sec:exp_details}.

\paragraph{Linear mode connectivity.}
To understand cross-lingual generalization through the lens of parameter space, we use linear mode connectivity.
LMC tests whether two models trained from the same initialization occupy the same error basin~\citep{frankle2020linear,neyshabur2020what,entezari2022the}.
We adapt it to the cross-lingual setting by measuring connectivity between an English adapter and each target-language adapter, both fine-tuned from the same base model.
Given English adapter parameters $\boldsymbol{\theta}_\text{en}$ and target-language adapter parameters $\boldsymbol{\theta}_\text{tgt}$, we evaluate the task metric $M(\boldsymbol{\theta}_\alpha)$ along the linear path $\boldsymbol{\theta}_\alpha = (1 - \alpha)\boldsymbol{\theta}_\text{en} + \alpha\boldsymbol{\theta}_\text{tgt}$ for $\alpha \in \{0, 0.1, \ldots, 1.0\}$.
If performance does not degrade along this path, no error barrier exists between the two adapters, indicating that they are linearly connected and lie in the same error basin.
Specifically, we regard a adapter pair as linearly connected when no valley appears along the path and the English adapter matches the target-language adapter in performance.
The barrier height~\citep{entezari2022the} quantifies the maximum performance drop along this path:
\begin{equation}
  \label{eq:barrier_height}
  B = \sup_\alpha [(1 - \alpha)M(\boldsymbol{\theta}_\text{en}) + \alpha M(\boldsymbol{\theta}_\text{tgt}) - M(\boldsymbol{\theta}_\alpha)].
\end{equation}

Since the target-language adapter is fine-tuned directly on target-language task data, it converges to a low-error basin on the target-language distribution.
When the English adapter is linearly connected to it, the interpolation path remains within this basin, placing the English adapter in the same target-language low-error region.
We therefore interpret LMC between English and target-language adapters as parameter-space evidence of cross-lingual generalization.
Specifically, the English adapter generalizes to the target-language distribution as effectively as the target-language oracle adapter.

\vspace{-0.3em}
\paragraph{Multilingual workflow.}
The internal behavior of multilingual LLMs has been characterized by the multilingual workflow (MWork) hypothesis, under which a multilingual LLM processes an input through three sequential stages~\citep{zhao2024how, wendler2024llamas}.
The early-layer stage encodes input tokens, written in the query language, into a concept-space representation that is largely language-agnostic (and empirically closer to English).
The middle-layer stage operates on this representation to build a semantically correct next-token representation.
The final-layer stage then decodes this representation back into the output space of the original input language.
We utilize MWork to interpret LMC between an English adapter and a target-language adapter.
When the two adapters are not linearly connected, we attribute the failure to the English adapter lacking sufficient ability at one or more MWork stages.
Building on this view, we perform a layer-wise analysis to localize where such deficiency arises.
\vspace{-3pt}

{\color{black}{\section{Cross-lingual generalization is mirrored by linear mode connectivity}\label{sec:R1}}}
\begin{figure*}[ht!]
  \centering
  \includegraphics[width=\linewidth]{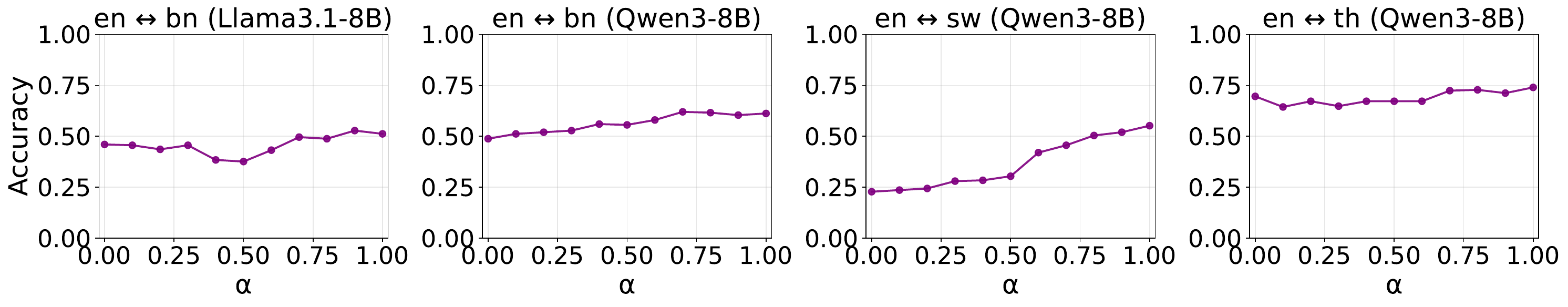}
  \caption{
    Linear interpolation path between English adapter $(\alpha = 0)$ and target-language adapter $(\alpha = 1)$ on Math.
    Each subplot shows accuracy evaluated on MGSM in the respective target language.
  }
  \label{fig:lmc_math}
  \vspace{-7pt}
\end{figure*}

\paragraph{Summary.}
Cross-lingual generalization manifests as LMC in parameter space between the English and target-language adapters.
This connectivity is sharply task-dependent.
It holds for NLI and QA but breaks for Math.
Prior work has offered limited insight into why it varies in this way, treating cross-lingual generalization as a largely monolithic phenomenon.

\subsection{NLI and QA}
\paragraph{Setup.}
For NLI, we train on the machine-translated multilingual NLI dataset~\citep{laurer2024less} and evaluate on XNLI~\citep{conneau2018xnli}, covering Arabic (\texttt{ar}), German (\texttt{de}), Spanish (\texttt{es}), French (\texttt{fr}), Hindi (\texttt{hi}), Russian (\texttt{ru}), Swahili (\texttt{sw}), Vietnamese (\texttt{vi}), and Chinese (\texttt{zh}).
For QA, we train on the translated SQuAD data from XTREME~\citep{hu2020xtreme} and evaluate on XQuAD~\citep{artetxe2020on}, covering Arabic (\texttt{ar}), German (\texttt{de}), Spanish (\texttt{es}), Hindi (\texttt{hi}), Russian (\texttt{ru}), Thai (\texttt{th}), Vietnamese (\texttt{vi}), and Chinese (\texttt{zh}).
These languages were chosen to cover commonly used low-, medium-, and high-resource languages.
In the main text, we report Spanish (\texttt{es}), Hindi (\texttt{hi}), and Chinese (\texttt{zh}) for both tasks, as well as Swahili (\texttt{sw}) for NLI and Thai (\texttt{th}) for QA, with the remaining languages provided in Appendix~\ref{appdix_sec:add_results}.
Task metrics are accuracy for NLI and F1 for QA.

\paragraph{NLU exhibits no error barrier and satisfies cross-lingual generalization.}
The NLI and QA paths in Figure~\ref{fig:lmc_main} are mostly flat with no error barrier.
For these NLU tasks, the English and target-language adapters lie in the same low-error basin, indicating that the English adapter already generalizes to the target language.
As edge cases, a single NLI pair (\texttt{sw}) and a single QA pair (\texttt{th}) show a slight rise in the middle of the interpolation path (third column in Figure~\ref{fig:lmc_main}).
This mild fluctuation still keeps both adapters within a single shared basin on the target language.

\subsection{Mathematical reasoning}
\label{sec:lmc_math}
\paragraph{Setup.}
For mathematical reasoning (Math), we train on MGSM8KInstruct~\citep{chen2024breaking} and evaluate on MGSM~\citep{shi2023language}.
MGSM covers Bengali (\texttt{bn}), German (\texttt{de}), Spanish (\texttt{es}), French (\texttt{fr}), Japanese (\texttt{ja}), Russian (\texttt{ru}), Swahili (\texttt{sw}), Thai (\texttt{th}), and Chinese (\texttt{zh}), from which we retain a model-specific subset (Table~\ref{tab:lang-selection}).
We restrict our analysis to (model, language) pairs whose target-language adapter improves over the zero-shot instruction-tuned model (the base model in our setup) on the LMC evaluation set.
This improvement ensures that the target-language adapter is the oracle that our analysis assumes, rather than a failed one.
Such failures occur when the instruction-tuned model is already strong and small-scale fine-tuning degrades it below the zero-shot baseline.
For NLI and QA, every language satisfies the criterion for every model, so all models share the same language set in each task.
For Math, the screen is selective and produces model-specific sets, reported in Table~\ref{tab:lang-selection}.
Task metrics are accuracy for Math.

\begin{table}[t]
\centering
\footnotesize
\resizebox{0.48\textwidth}{!}{%
\begin{tabular}{c|ccc}
\hline
Lang & Llama3.1-8B & Qwen3-8B & Gemma2-9B \\
\hline
\texttt{bn} & \textbf{+0.190} & \textbf{+0.296} & -0.072 \\
\texttt{de} & -0.140 & -0.064 & -0.068 \\
\texttt{es} & -0.108 & -0.056 & -0.104 \\
\texttt{fr} & -0.156 & -0.108 & -0.168 \\
\texttt{ja} & -0.088 & -0.152 & -0.100 \\
\texttt{ru} & -0.132 & -0.044 & -0.116 \\
\texttt{sw} & -0.104 & \textbf{+0.080} & -0.048 \\
\texttt{th} & -0.112 & \textbf{+0.108} & -0.060 \\
\texttt{zh} & -0.132 & -0.128 & -0.052 \\
\hline
\end{tabular}%
}
\caption{Language selection for Math (MGSM).
Each entry is the accuracy improvement of the target-language adapter over the base model (i.e., instruction-tuned model) on the target-language test set.
Entries with positive value (selected pairs) are shown in bold.}
\label{tab:lang-selection}
\vspace{-7pt}
\end{table}

\paragraph{Math adapters are not linearly connected and fail to satisfy cross-lingual generalization.}
Figure~\ref{fig:lmc_math} shows that in Math, the English and target-language adapters consistently fail to share a basin on the target language, although the form of this failure varies across pairs.
First, for Llama3.1-8B on \texttt{bn} and Qwen3-8B on \texttt{th}, an error barrier separates the two adapters, placing them in different basins even though their endpoint performance is comparable.
This leaves it unclear whether the English adapter generalizes to the target language.
The monotonically increasing cases also occur for Qwen3-8B on \texttt{bn} and \texttt{sw}, with sharper rises than in the NLU setting, most pronounced for \texttt{sw}.
This indicates that the English adapter lies even further outside the target-language basin and thus lacks, to an even greater extent than in NLU, the additional cross-lingual ability that Math requires.
In summary, the contrast between NLU and Math shows that cross-lingual generalization is not monolithic.
Its basin geometry is governed by task type, with most pairs linearly connected in NLU and LMC rarely observed in Math.

{\color{black}{\section{Localizing the abilities that improve cross-lingual generalization}\label{sec:R2}}}
\begin{figure*}[ht]
  \centering
  \includegraphics[width=\linewidth]{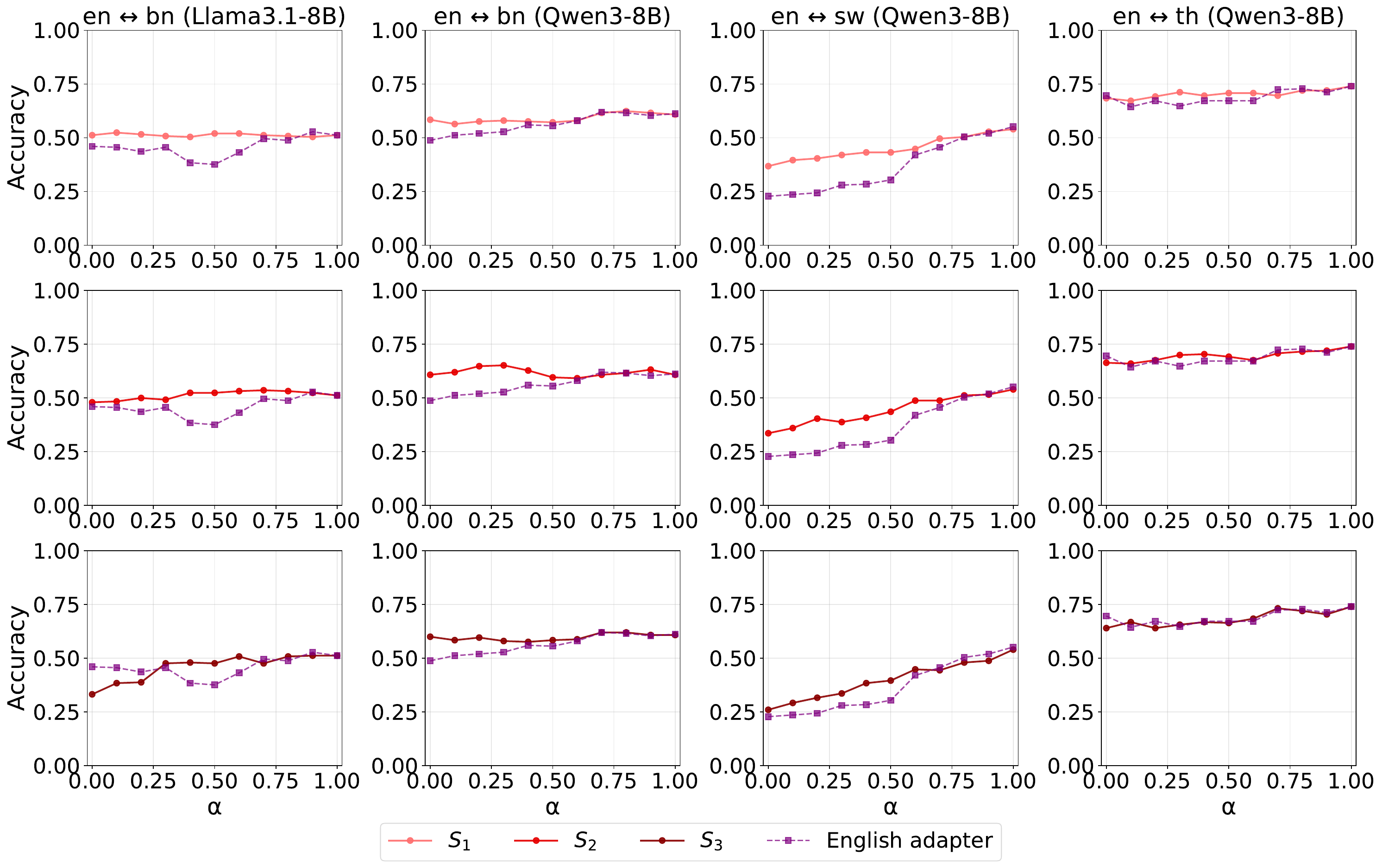}
  \caption{
    Linear interpolation path between ability-adapted English adapter $(\alpha = 0)$ and target-language adapter $(\alpha = 1)$ on Math.
    Rows correspond to early-layer ($S_1$), middle-layer ($S_2$), and final-layer ($S_3$) adaptation, respectively.
    Each column corresponds to a target language, and each subplot shows accuracy evaluated on MGSM in that language.
  }
  \label{fig:con_lmc}
  \vspace{-15pt}
\end{figure*}

\begin{figure*}[ht]
  \centering
  \includegraphics[width=\linewidth]{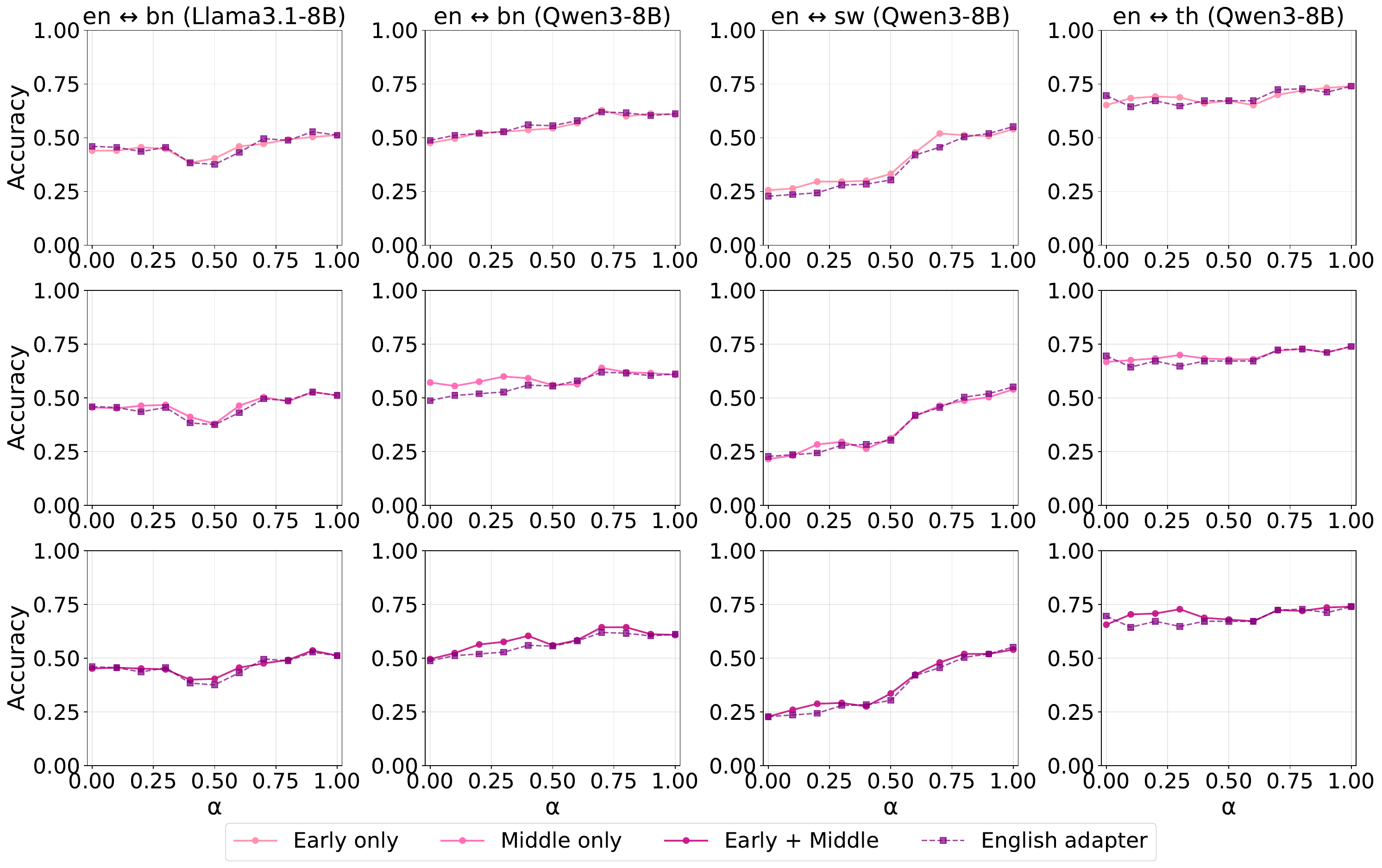}
  \caption{
    Linear interpolation path between routed-adaptation English adapter $\tilde{\boldsymbol{\theta}}_\text{en} $ $(\alpha=0)$ and target-language adapter $\boldsymbol{\theta}_\text{tgt} $ $(\alpha=1)$ on Math.
    Rows correspond to routing only the corpus signal to $S_1$ (Early only), routing only the English task signal to $S_2$ (Middle only), and routing both signals jointly (Early and Middle), respectively.
    Each column corresponds to a target language, and each subplot shows accuracy evaluated on MGSM in that language.
  }
  \label{fig:routing_lmc}
\end{figure*}

\paragraph{Summary.}
When cross-lingual generalization breaks, the deficit is local rather than global, recoverable by adapting a single layer group and confined to the early and middle layers.
Section~\ref{sec:R1} showed that this breakdown occurs for Math while NLU pairs stay connected, which leaves open whether the Math gap reflects a global mismatch or a localized one.
Relative to the target-language oracle, we ask which ability the English adapter lacks on Math, and whether the deficit can be repaired by adapting only the corresponding layers.
We pursue this question through MWork, which assigns distinct functional roles to the early, middle, and final layers.
We localize the deficit by adapting each group in isolation on the target-language task dataset.
The group that restores connectivity then pinpoints the ability the English adapter lacks.
\vspace{3pt}

\subsection{Setup}
\paragraph{Layer-group adaptation.}
Recall that MWork attributes early layers to language-specific input encoding, middle layers to language-agnostic semantic composition, and final layers to language-specific output decoding.
We isolate each of these three layer groups and continually fine-tune them to enhance the corresponding ability.
We then examine how this targeted adaptation affects cross-lingual LMC and, in turn, cross-lingual generalization.
Specifically, for each target language, we randomly sample 500 target-language training examples from the Math dataset (MGSM8KInstruct).
Starting from the English adapter parameters $\boldsymbol{\theta}_\text{en}$, we continually fine-tune only a chosen layer group $S \subseteq \{1, \ldots, L\}$ on these samples for 3 epochs.
The remaining layers stay frozen at the English adapter values.
We then evaluate the task metric $M(\boldsymbol{\theta}_\alpha^S)$ along the linear path $\boldsymbol{\theta}_\alpha^S = (1 - \alpha)\boldsymbol{\theta}_\text{en}^S + \alpha\boldsymbol{\theta}_\text{tgt}$ between the ability-adapted English adapter $\boldsymbol{\theta}_\text{en}^S$ and the target-language adapter $\boldsymbol{\theta}_\text{tgt}$, with $\alpha \in \{0, 0.1, \ldots, 1.0\}$.
A previously disconnected pair may become linearly connected once we strengthen a specific ability.
We take this as evidence that the corresponding ability is what the English adapter lacks for cross-lingual generalization.

\paragraph{Analysis setup.}
We report results across all three layer groups: early ($S_1 = \{1, \ldots, \lceil L/3 \rceil \}$), middle ($S_2 = \{\lceil L/3 \rceil + 1, \ldots, \lceil 2L/3 \rceil \}$), and final layers ($S_3 = \{\lceil 2L/3 \rceil + 1, \ldots, L \}$).
We focus this analysis on Math, the task whose adapter pairs are linearly disconnected.

\subsection{Results}
\paragraph{One of the abilities the English adapter lacked is language-specific input encoding.}
Adapting only the early layers makes most adapter pairs linearly connected on Math (first row of Figure~\ref{fig:con_lmc}).
Early-layer adaptation can relocate the English adapter into the target-language basin without altering the remaining layers.
The only residual case is Qwen3-8B on \texttt{sw}, which does not become linearly connected, yet its steep monotonic increase before adaptation is clearly attenuated.
Within MWork, this restoration indicates that the English adapter's input encoding generalizes less well to target-language inputs than the target-language adapter.
Early-layer adaptation thus identifies language-specific input encoding as one of the abilities the English adapter must improve for cross-lingual generalization.

\paragraph{Language-agnostic semantic composition is another ability in which the English adapter was deficient.}
Middle-layer adaptation also makes most adapter pairs linearly connected on Math, restoring the English adapter's semantic composition to the level of the target-language adapter (second row of Figure~\ref{fig:con_lmc}).
Qwen3-8B on \texttt{sw} still increases monotonically, but its barrier is mitigated relative to the naive English adapter.
This suggests that the naive English adapter's language-agnostic semantic composition does not fully generalize to target-language inputs as well as the target-language adapter does.
Middle-layer adaptation thus identifies language-agnostic semantic composition as another ability the English adapter lacks relative to the target-language adapter, which can limit cross-lingual generalization.

\paragraph{Language-specific output decoding is less beneficial for achieving cross-lingual generalization.}
Among the three layer groups, final-layer adaptation produces the weakest connectivity, although only some of its error barriers become milder after adaptation (third row of Figure~\ref{fig:con_lmc}).
More pairs fail to become linearly connected than under early- or middle-layer adaptation, and these failures show increases, with Llama3.1-8B on \texttt{bn} as the most notable case.
The final layers likely still operate on the original upstream English representation, so adapting their decoding alone may not relocate the adapter into the target-language basin.
Final-layer adaptation thus identifies language-specific decoding as an ability that does not drive cross-lingual generalization, in contrast to the early and middle layers.

{\color{black}{\section{Factorizing cross-lingual generalization without target-language labels}\label{sec:R3}}}
\paragraph{Summary.}
Cross-lingual generalization factorizes in parameter space into two components, a language-specific input encoding and a language-agnostic semantic composition.
Each can be supplied separately from a cheaper signal matched to its role under MWork, without any target-language task labels.
The localization in Section~\ref{sec:R2} relied on the full target-language task signal, which bundles language and task together.
We establish factorization by decomposing this signal and routing each part to its matched layers.
We route a target-language corpus to $S_1$, since input encoding is language-specific.
Semantic composition, by contrast, is language-agnostic, so English task data alone suffices at $S_2$.

\vspace{5pt}

\subsection{Setup}
\paragraph{Routed adaptation.}
Following the per-layer adaptation of Section~\ref{sec:R2}, we now route two distinct training signals to two disjoint layer groups.
Specifically, for each target language we randomly sample 500 sentences from the target-language CulturaX~\citep{nguyen2024culturax} corpus and 500 English training examples from the Math dataset (MGSM8KInstruct).
Starting from the English adapter parameters $\boldsymbol{\theta}_\text{en}$, we jointly fine-tune the language encoding layers $S_1$ on the corpus and the composition layers $S_2$ on the English task for 3 epochs.
This yields a routed English adapter $\tilde{\boldsymbol{\theta}}_\text{en}$.
We then evaluate the task metric $M(\tilde{\boldsymbol{\theta}}_\alpha)$ along the linear path $\tilde{\boldsymbol{\theta}}_\alpha = (1-\alpha)\tilde{\boldsymbol{\theta}}_\text{en} + \alpha\boldsymbol{\theta}_\text{tgt}$ for $\alpha = \{0, 0.1, \ldots, 1.0\}$.

\paragraph{Analysis setup.}
We compare three routing variants, namely the corpus signal alone to $S_1$  (Early only), the English task signal alone to $S_2$ (Middle only), and both signals jointly (Early and Middle).
Throughout this section, we quantify the effect of routed adaptation as the reduction in barrier height $B$ (Eq.~\ref{eq:barrier_height}) relative to the naive English adapter.

\subsection{Results}
\paragraph{Semantic composition is the most dominant factor.}
Each single-signal routing lowers the barrier on Math for nearly every pair, without any target-language task labels (second and third rows of Table~\ref{tab:barrier}, first and second rows of Figure~\ref{fig:routing_lmc}).
Composition routing (Middle only), the English task at $S_2$, is the stronger factor.
On average it lowers the barrier from 0.102 to 0.084 and is the only variant to raise accuracy above the naive adapter, from 0.468 to 0.478.
Language routing (Early only), the corpus at $S_1$, also lowers the barrier, to 0.086, but slightly reduces accuracy.
Cross-lingual generalization thus factorizes into a language-specific input encoding and a language-agnostic semantic composition, with semantic composition the most important factor.

\paragraph{Composition alone accounts for much of the gain.}
Adding the language signal to composition does not improve on composition alone (fourth row of Table~\ref{tab:barrier}, third row of Figure~\ref{fig:routing_lmc}).
The joint routing (Early and Middle) matches composition routing on the average barrier (0.084) but reaches a lower accuracy of 0.458, below the latter's 0.478.
Composition routing therefore attains the highest accuracy and the lowest barrier of any variant.
This result is also shown for Qwen3-8B \texttt{bn}, where composition routing reaches the lowest barrier along its interpolation path (Figure~\ref{fig:routing_lmc}).
The other routings lower the barrier less than composition alone.
Composition thus emerges as the primary, label-free driver of cross-lingual generalization.

\vspace{10pt}

\begin{table}[ht]
\centering
\footnotesize
\resizebox{0.3\textwidth}{!}{%
\begin{tabular}{l|c}
\hline
Baseline & Avg Acc. (Avg. $B$) \\
\hline
naive & 0.468 (0.102) \\
\hline
E & 0.456 (0.086) \\
M & \textbf{0.478} (\textbf{0.084}) \\
E+M & 0.458 (\textbf{0.084}) \\
\hline
\end{tabular}%
}
\caption{Average accuracy and barrier height $B$ of the routing variants on Math (MGSM).
Averages are computed over the four (model, language) pairs: Llama3.1-8B bn, Qwen3-8B bn, Qwen3-8B sw, and Qwen3-8B th.
Rows denote the naive English adapter (naive) and three routing variants, namely Early only (E), Middle only (M), and Early and Middle (E+M).
\textbf{Bold} denotes the highest accuracy and lowest barrier height.}
\label{tab:barrier}
\vspace{-10pt}
\end{table}

\paragraph{Composition routing extends an English model to a new language without target-language data.}
This factorization has a concrete practical consequence.
Because the composition layers are language-agnostic, English task supervision at $S_2$ alone improves cross-lingual connectivity for nearly every pair.
This requires no target-language data, neither task labels nor an unlabeled corpus.
The supervision also uses only a small number of English task samples, which keeps the recipe practical in low-resource settings.
Target-language task annotations, by contrast, remain a primary bottleneck for cross-lingual adaptation.
Composition routing thus offers a label-efficient path to extending an English-trained model to a new target language.

\section{Conclusion}
Cross-lingual generalization is governed by whether English and target-language adapters share a low-error basin in parameter space.
Using linear mode connectivity (LMC) between English and target-language LoRA adapters across three tasks, we surface three findings.
First, the geometry is task-dependent, with NLI and QA model-language pairs mostly linearly connected while mathematical reasoning (Math) pairs largely fail to connect.
Second, the Math deficit is layer-localized, since a small target-language task sample on only the language encoding or semantic composition layers can restore LMC.
Third, the geometry is factorizable.
An unlabeled target-language corpus routed to the language-specific layers and English task supervision routed to the composition layers each lower the barrier on their own for nearly every pair.
Composition is the stronger of the two, as English supervision on it alone achieves this without any target-language data, offering a practical route for low-resource cross-lingual adaptation.

\section*{Limitations}
Our study has several limitations. 
First, mathematical reasoning is the only non-NLU task in our experiments.
Our experimental design requires parallel multilingual training data translated from the same English source, so that differences in basin structure can be attributed to language rather than data content.
Few multilingual reasoning datasets satisfy this constraint.
Second, MWork was originally formulated based on observations of pre-trained models, and its layer-wise functional decomposition may not transfer cleanly to fine-tuned adapters.
Our interpretations under MWork should therefore be regarded as suggestive rather than conclusive.
Third, our experiments are limited to 8B-parameter models, and whether larger or smaller models exhibit the same basin structure has not been tested.
Fourth, due to a limited computational budget, our analysis is confined to a LoRA-only setting.
Whether full fine-tuning yields the same connectivity patterns remains an open question.
Fifth, our Math analysis covers only the small and model-specific subset of (model, language) pairs that pass the adapter-improvement criterion, leaving most languages and the entire Gemma model outside its scope.
With broader multilingual reasoning datasets, future work could extend the analysis to the languages and models that this criterion currently excludes.

{\color{black}{\section*{Acknowledgements}
This work was supported by Institute of Information \& communications Technology Planning \& Evaluation (IITP) grant funded by the Korea government(MSIT) [NO.RS-2021-II211343, Artificial Intelligence Graduate School Program (Seoul National University)] and Institute of Information \& communications Technology Planning \& Evaluation (IITP) grant funded by the Korea government (MSIT) (No. RS-2026-25522672, Development of Unified Reasoning Technology Mimicking Human Cognition for Hierarchical Understanding and Unbounded Problem Solving).}}

\bibliography{custom}

\appendix

\section{Experimental details}
\label{appdix_sec:exp_details}

\subsection{Datasets}
The per-language sizes of our training and evaluation sets vary across tasks.
For training, we draw on the machine-translated multilingual NLI set with approximately 25,000 examples per language for NLI~\citep{laurer2024less}, the translated SQuAD set with approximately 80,000 for QA~\citep{hu2020xtreme}, and MGSM8KInstruct~\citep{chen2024breaking} with approximately 7,500 for Math.
From each set, we sample 5,000 examples per language so that the training size is fixed across tasks.
For evaluation, we test NLI on the XNLI test set with 5,000 examples per language~\citep{conneau2018xnli}, QA on the XQuAD test set with 1,190~\citep{artetxe2020on}, and Math on MGSM~\citep{shi2023language} with 250.

\subsection{Training}
All adapters use LoRA~\citep{hu2022lora} with rank 16, alpha 32, and dropout 0.05, applied to the query, key, value, and output projections as well as the gate, up, and down projections.
The maximum sequence length is set per task to match the input distributions, with 1,024 tokens for Math, 512 for NLI, and 1,536 for QA.
These lengths reflect the long reasoning chains in MGSM8KInstruct, the short premise and hypothesis pairs in NLI, and the Wikipedia-style passages in QA.
All adapters are trained by using 4 NVIDIA RTX A6000 GPUs.

We optimize with a cosine schedule, a warmup ratio of 0.15, and weight decay of 0.01.
The learning rate is $10^{-4}$ for Math, $2\times10^{-4}$ for NLI, and $10^{-4}$ for QA.
We train all three tasks for 3 epochs.
The effective batch size is held at 16 across tasks, with the per-device batch size and gradient accumulation steps set to 2 and 8 for Math and QA and to 4 and 4 for NLI.
Training uses bfloat16 mixed precision, gradient clipping at 0.5, and a fixed seed of 42.

\begin{table*}[t]
\centering
\footnotesize
\begin{tabular}{l|cccc|c}
\hline
Baseline & Llama3.1-8B \texttt{bn} & Qwen3-8B \texttt{bn} & Qwen3-8B \texttt{sw} & Qwen3-8B \texttt{th} & Avg Acc. (Avg. $B$) \\
\hline
naive & \textbf{0.460} (0.110) & 0.488 (0.062) & 0.228 (0.162) & \textbf{0.696} (0.074) & 0.468 (0.102) \\
\hline
E & 0.440 (0.092) & 0.476 (0.066) & \textbf{0.256} (\textbf{0.142}) & 0.652 (0.044) & 0.456 (0.086) \\
M & 0.456 (0.104) & \textbf{0.572} (\textbf{0.034}) & 	0.216 (0.162) & 0.668 (\textbf{0.036}) & \textbf{0.478} (\textbf{0.084}) \\
E+M & 0.452 (\textbf{0.082}) & 0.496 (0.056) & 0.228 (0.156) & 0.656 (0.042) & 0.458 (\textbf{0.084}) \\
\hline
\end{tabular}
\caption{Per-pair accuracy and barrier height $B$ of the routing variants on Math (MGSM).
There are four baselines with three routing variants: naive English adapter (naive), namely Early only (E), Middle only (M), and Early and Middle (E+M).
For each model-language pair, \textbf{bold} denotes the highest accuracy and lowest barrier height.}
\label{tab:per_barrier}
\end{table*}

We consider the following adaptation regimes, all initialized from the English adapter:
\begin{itemize}
    \item \textbf{Ability-adapted English adapter} $\boldsymbol{\theta}_\text{en}^{S_k}$: From the English adapter $\boldsymbol{\theta}_\text{en}$, freeze all LoRA parameters outside layer group $S_k \in \{S_1, S_2, S_3\}$ and continue SFT on target-language task data.
    We use random-sampled 500 examples per language with seed 42.
    \item \textbf{Routed-adaptation English adapter} $\tilde{\boldsymbol{\theta}}_\text{en}$: From the English adapter, freeze all LoRA parameters outside the assigned routing slices and continue training with two parallel SFT losses.
    We use random-sampled 500 examples per language with seed 42.
    The total loss is defined as:
    $$\mathcal{L}_{\text{total}} = \lambda_{\text{lang}} \mathcal{L}_{\text{lang}} + \lambda_{\text{task}} \mathcal{L}_{\text{task}},$$
    where $\mathcal{L}_\text{lang}$ is the standard causal-LM SFT cross-entropy computed on the target-language CulturaX~\citep{nguyen2024culturax} corpus (SFT-formatted), and $\mathcal{L}_\text{task}$ is computed using only English task data.
    We set $\lambda_\text{lang} = \lambda_\text{task} = 1$.
    Each loss term exclusively updates the parameters of its assigned layer slice (i.e., $\mathcal{L}_{\text{lang}}$ updates layers $k \in S_{\text{lang}}$, and $\mathcal{L}_{\text{task}}$ updates layers $k \in S_{\text{task}}$).
    Specifically, the Language routing (Early only) variant isolates updates to $S_1$ via $\mathcal{L}_{\text{lang}}$, leaving $\mathcal{L}_{\text{task}}$ inactive ($S_{\text{task}} = \emptyset$). The Composition routing (Middle only) configuration does the reverse, updating $S_2$ via $\mathcal{L}_{\text{task}}$ while $\mathcal{L}_{\text{lang}}$ is disabled. In Joint routing (Early + Middle), both losses are employed simultaneously, with $\mathcal{L}_{\text{lang}}$ modifying $S_1$ and $\mathcal{L}_{\text{task}}$ modifying $S_2$.
\end{itemize}

\subsection{Inference}
At inference, decoding is greedy with temperature 0 and a batch size of 8.
We set the maximum number of new tokens to 512 for Math, 8 for the NLI label, and 64 for the QA short answer.
Inputs are formatted with the LlamaFactory template~\citep{zheng2024llamafactory}.
Every inference was conducted using 4 NVIDIA RTX A6000 GPUs.

\subsection{Discussion of artifacts}
All artifacts used in this research are publicly available under licenses permitting academic use.
For the datasets, XNLI and MGSM are licensed under CC BY-NC 4.0, XQuAD under CC BY-SA 4.0, and MGSM8KInstruct under the MIT License, while the CulturaX corpus follows mC4 and OSCAR licenses.
For the experimental codebase, we utilized LlamaFactory, which operates under the Apache License 2.0.
Regarding the models, Qwen is distributed under the Apache License 2.0, whereas Llama and Gemma are used under their custom research licenses.

\begin{table}[ht]
\centering
\footnotesize
\resizebox{0.3\textwidth}{!}{%
\begin{tabular}{c|cc}
\hline
Lang & MGSM & MSVAMP \\
\hline
\texttt{bn} & \textbf{+0.104} & \textbf{+0.022} \\
\texttt{de} & \textbf{+0.184} & -0.032 \\
\texttt{es} & \textbf{+0.060} & -0.073 \\
\texttt{fr} & \textbf{+0.068} & -0.060 \\
\texttt{ja} & \textbf{+0.004} & -0.023 \\
\texttt{ru} & \textbf{+0.116} & -0.056 \\
\texttt{sw} & \textbf{+0.268} & \textbf{+0.372} \\
\texttt{th} & \textbf{+0.080} & -0.040 \\
\texttt{zh} & -0.104 & -0.141 \\
\hline
\end{tabular}%
}
\caption{Language selection for Math (MGSM and MSVAMP) on Qwen3-1.7B.
Each entry is the accuracy gain of the target-language adapter over the base model (i.e., instruction-tuned model) on the target-language test set.
\textbf{Bold} entries indicates pairs with positive gain (i.e., the selected pairs).}
\label{tab:lang-selection_1b}
\end{table}

\section{Additional results}
\label{appdix_sec:add_results}

\subsection{NLI and QA}
\begin{figure*}[ht]
  \centering
  \includegraphics[width=0.8\linewidth]{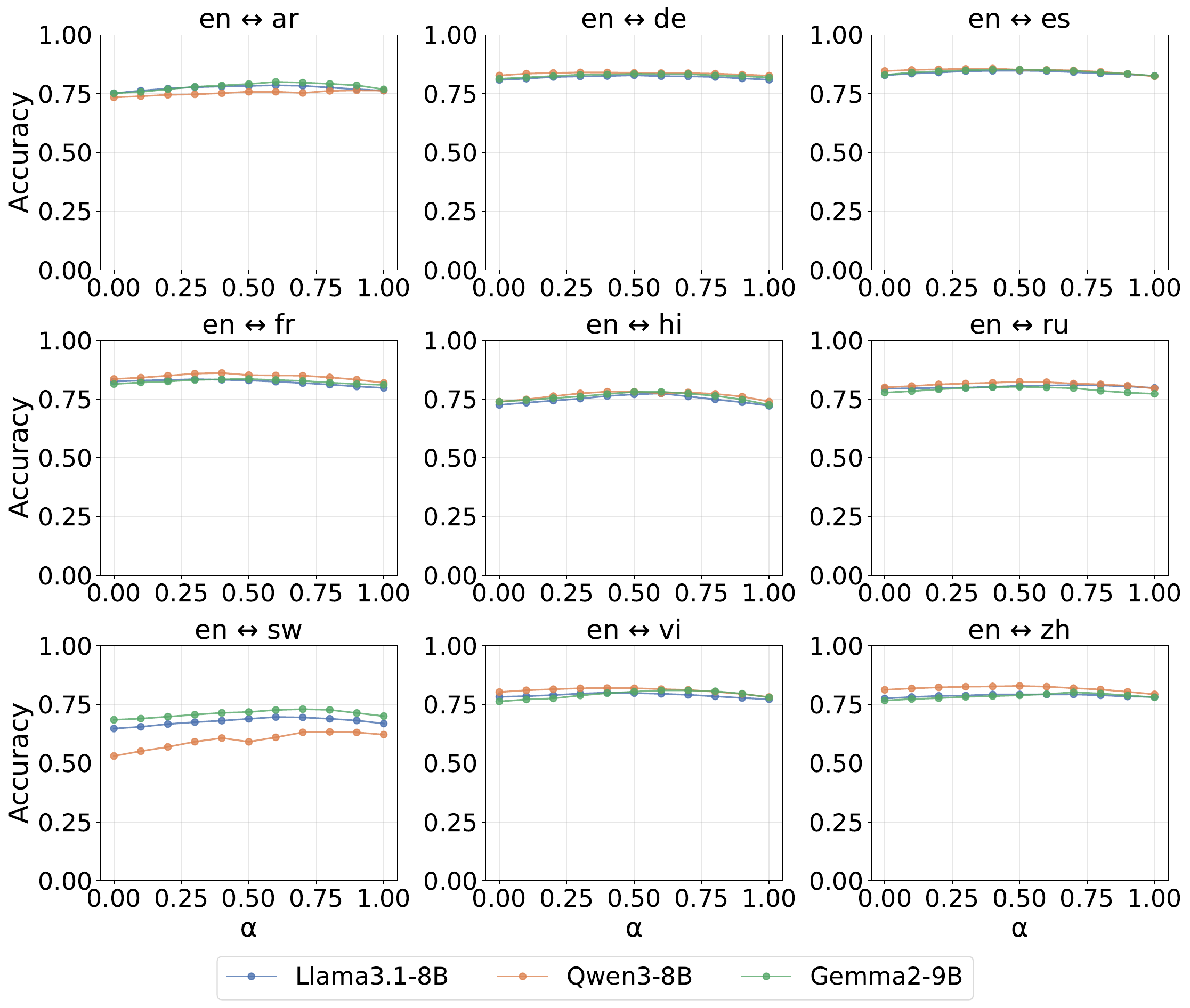}
  \caption{
    Linear interpolation path between English adapter $(\alpha = 0)$ and target-language adapter $(\alpha = 1)$ on NLI.
    Each subplot shows accuracy evaluated on XNLI in the respective target language.
  }
  \label{fig:lmc_nli}
\end{figure*}

\begin{figure*}[ht!]
\centering
\includegraphics[width=\linewidth]{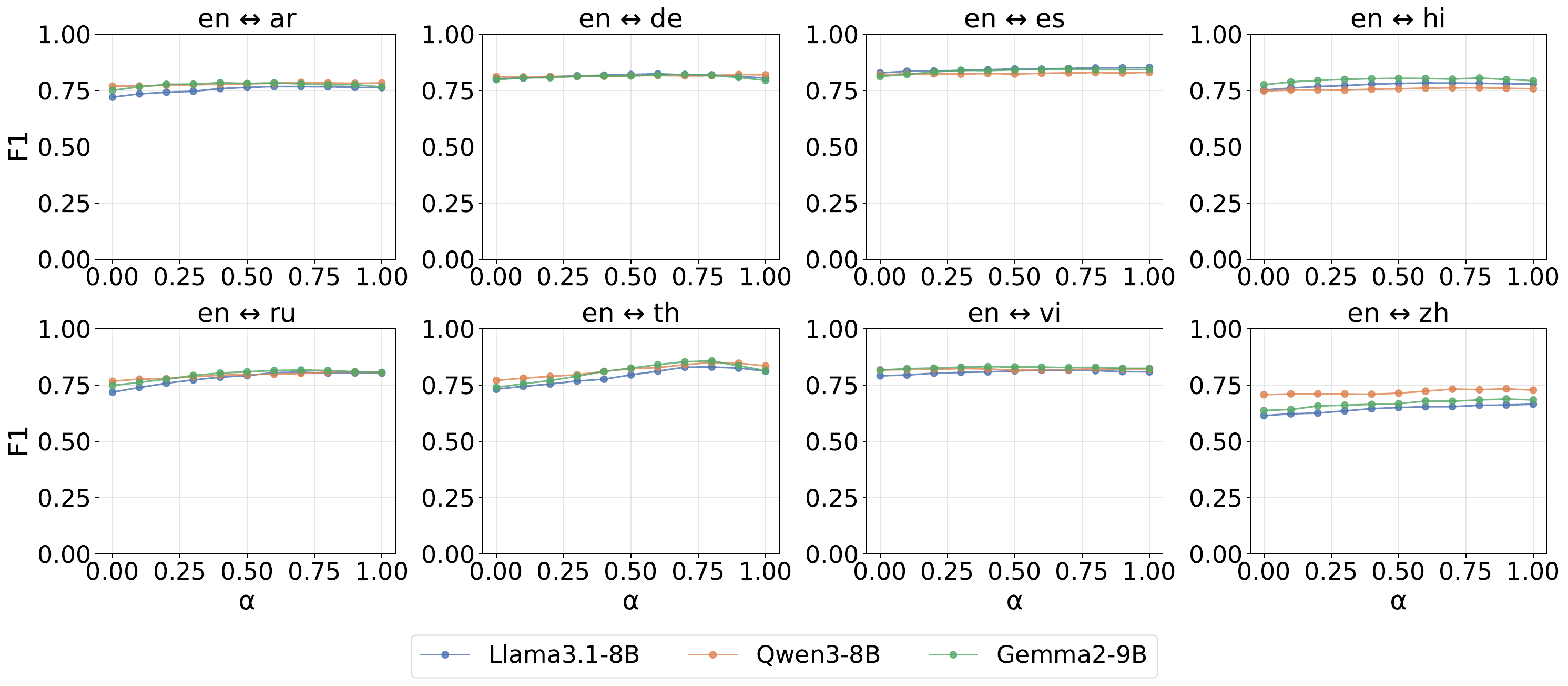}
  \caption{
    Linear interpolation path between English adapter $(\alpha = 0)$ and target-language adapter $(\alpha = 1)$ on QA.
    Each subplot shows F1 score evaluated on XQuAD in the respective target language.
  }
  \label{fig:lmc_qa}
\end{figure*}

The NLI and QA paths in Figures~\ref{fig:lmc_nli} and~\ref{fig:lmc_qa} are mostly flat with no error barrier.
For NLU tasks, the English and target-language adapters lie in the same low-error basin, indicating that the English adapter already generalizes to the target language.
As edge cases, three pairs rise monotonically from the English endpoint to the target-language endpoint: \texttt{sw} on NLI and \texttt{ru} and \texttt{th} on QA.
The first two appear in the main text, while \texttt{ru} is additionally surfaced here.
This indicates that the English adapter lies outside the target language's low-error basin, suggesting that it lacks, relative to the target-language adapter, the ability required for cross-lingual generalization.

\begin{table}[ht]
\centering
\footnotesize
\resizebox{0.3\textwidth}{!}{%
\begin{tabular}{l|c}
\hline
Baseline & Avg Acc. (Avg. $B$) \\
\hline
naive & 0.367 (0.029) \\
\hline
E & 0.276 (\textbf{0.014}) \\
M & \textbf{0.379} (0.023) \\
E+M & 0.330 (0.028) \\
\hline
\end{tabular}%
}
\caption{Average accuracy and barrier height $B$ of the routing variants of Qwen3-1.7B on Math (MGSM).
Averages are computed over 8 languages.
Rows denote the naive English adapter (naive) and three routing variants, namely Early only (E), Middle only (M), and Early and Middle (E+M).
\textbf{Bold} denotes the highest accuracy and lowest barrier height.}
\label{tab:barrier_1b}
\end{table}

\subsection{Sensitivity analysis of layer grouping}

\begin{figure*}[ht]
  \centering
  \includegraphics[width=\linewidth]{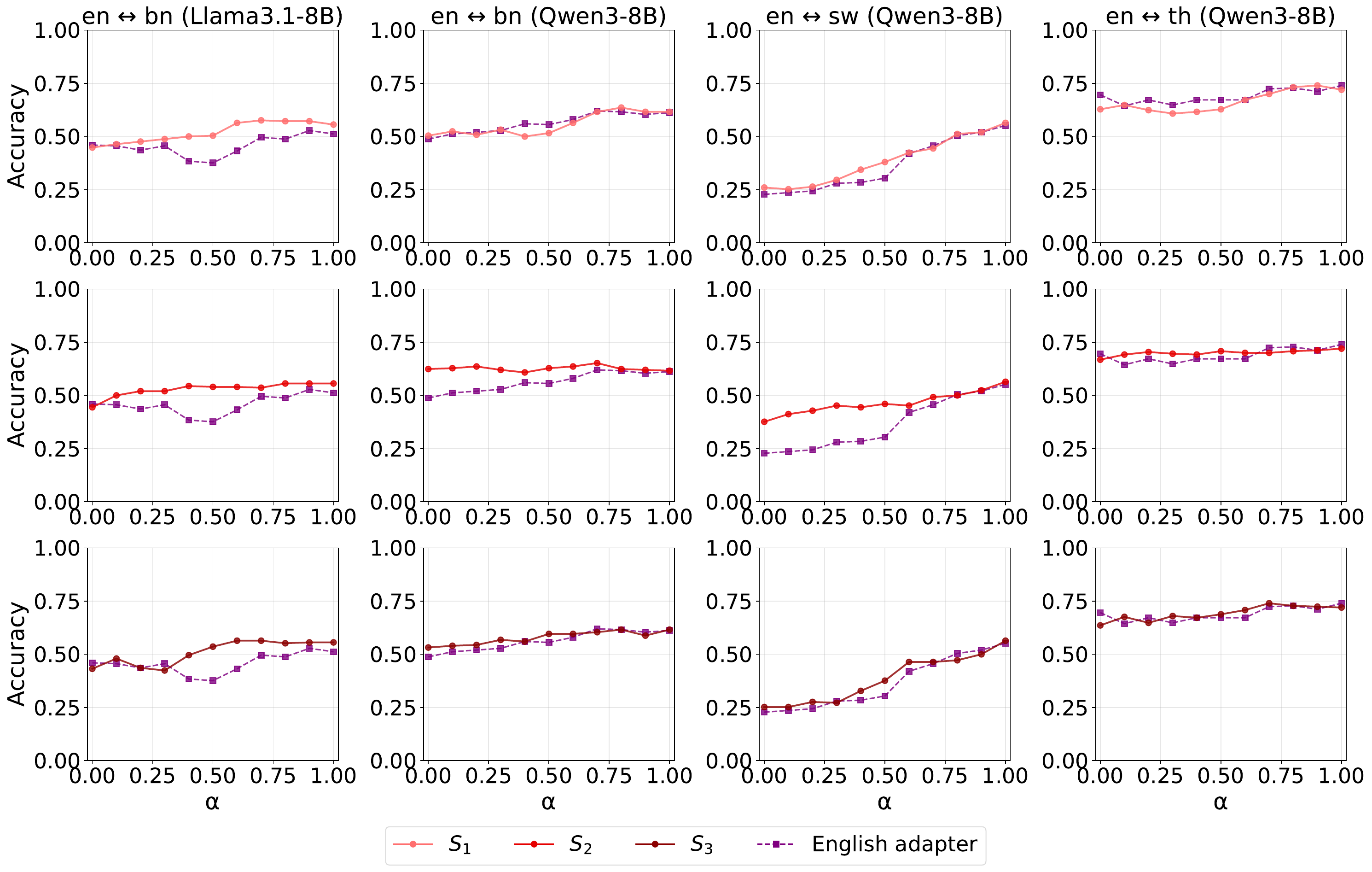}
  \caption{
    Linear interpolation path between ability-adapted English adapter $(\alpha = 0)$ and target-language adapter $(\alpha = 1)$ on Math.
    Rows correspond to early-layer ($S_1$), middle-layer ($S_2$), and final-layer ($S_3$) adaptation, respectively, where the layers are split in a $15\!:\!70\!:\!15$ ratio.
    Each column corresponds to a target language, and each subplot shows accuracy evaluated on MGSM in that language.
  }
  \label{fig:con_lmc_15}
\end{figure*}

\begin{figure*}[ht]
  \centering
  \includegraphics[width=\linewidth]{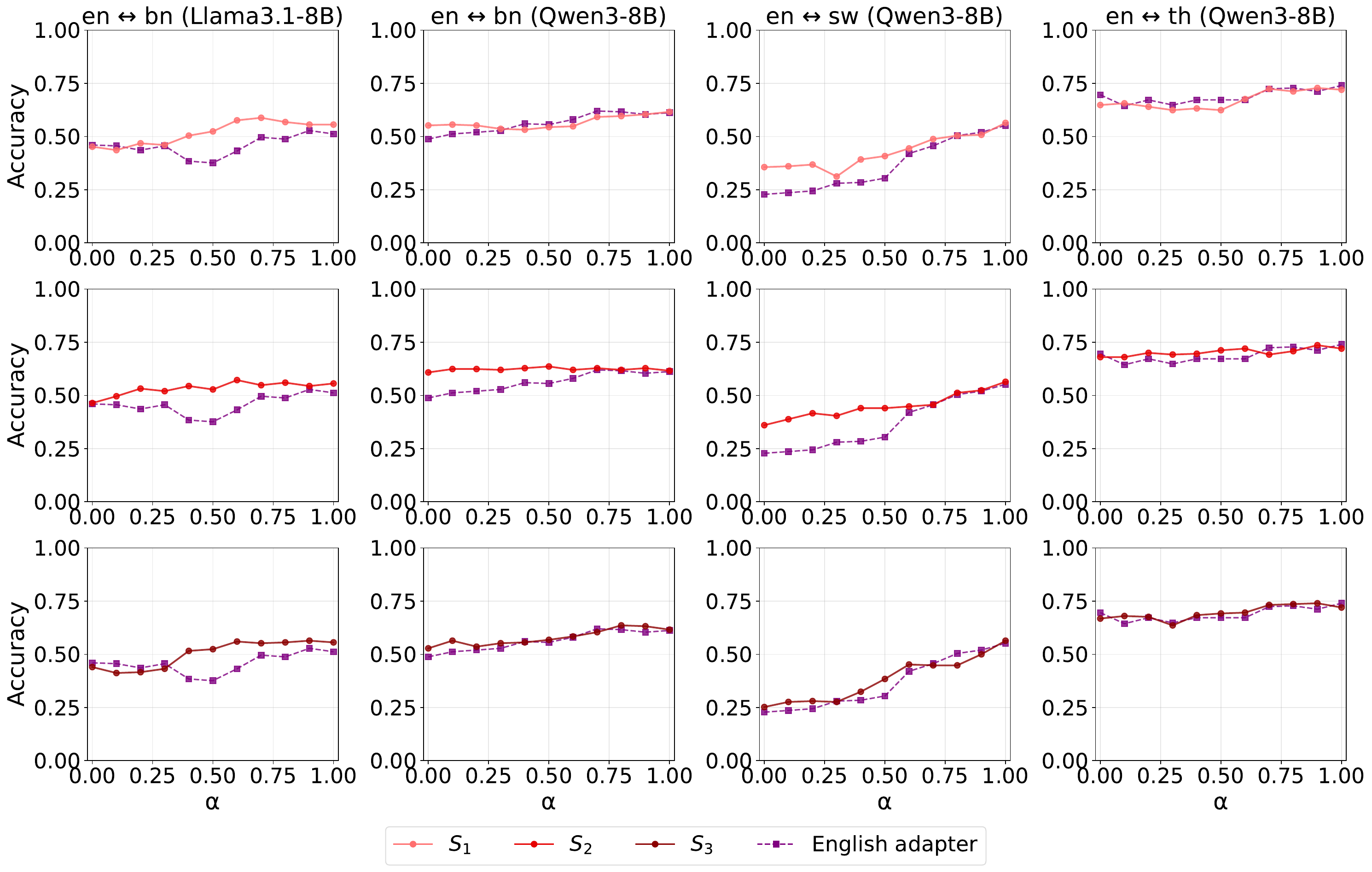}
  \caption{
    Linear interpolation path between ability-adapted English adapter $(\alpha = 0)$ and target-language adapter $(\alpha = 1)$ on Math.
    Rows correspond to early-layer ($S_1$), middle-layer ($S_2$), and final-layer ($S_3$) adaptation, respectively, where the layers are split in a $20\!:\!60\!:\!20$ ratio.
    Each column corresponds to a target language, and each subplot shows accuracy evaluated on MGSM in that language.
  }
  \label{fig:con_lmc_20}
\end{figure*}

We examine layer-group adaptation under two additional splits to verify that the layer-localized deficit in Section~\ref{sec:R2} holds even if the layer split is different.
Following prior work that treats the middle layer as performing language-agnostic composition~\citep{zhao2024how,bandarkar2025layer}, we split layers by expanding the middle layer group.
Specifically, we set two split variants: a $15:70:15$ and a $20:60:20$ split for the early ($S_1$), middle ($S_2$), and final ($S_3$) layer groups, respectively.

In Figure~\ref{fig:con_lmc_15} and Figure~\ref{fig:con_lmc_20}, layer-group adaptation typically makes adapters linearly connected regardless of the split, consistent with the main text.
Middle layer adaptation mitigates the barrier robustly across both splits, while early-layer adaptation becomes less effective as the middle-layer group grows larger.
This suggests that within the early layer group, layers closer to the middle layers contribute more to cross-lingual generalization.

\subsection{Per-pair breakdown of routing variants}

Table~\ref{tab:per_barrier} reports the barrier and accuracy of every model-language pair, complementing the average results (Table~\ref{tab:barrier}) given in the main text.
In most cases, every routing variant reduces the barrier compared to the English adapter.
We note that composition routing (Middle only) lowers the barrier height to about half that of the naive English adapter on Qwen3-8B \texttt{bn} and Qwen3-8B \texttt{th}.
Composition routing also yields a large accuracy improvement, from 0.488 to 0.572 on Qwen3-8B \texttt{bn}, and there is no destructive degradation on other pairs.

\section{Analysis on Qwen3-1.7B}
\label{appdix_sec:qwen3-1.7b}

\begin{figure*}[ht]
  \centering
  \includegraphics[width=0.7\linewidth]{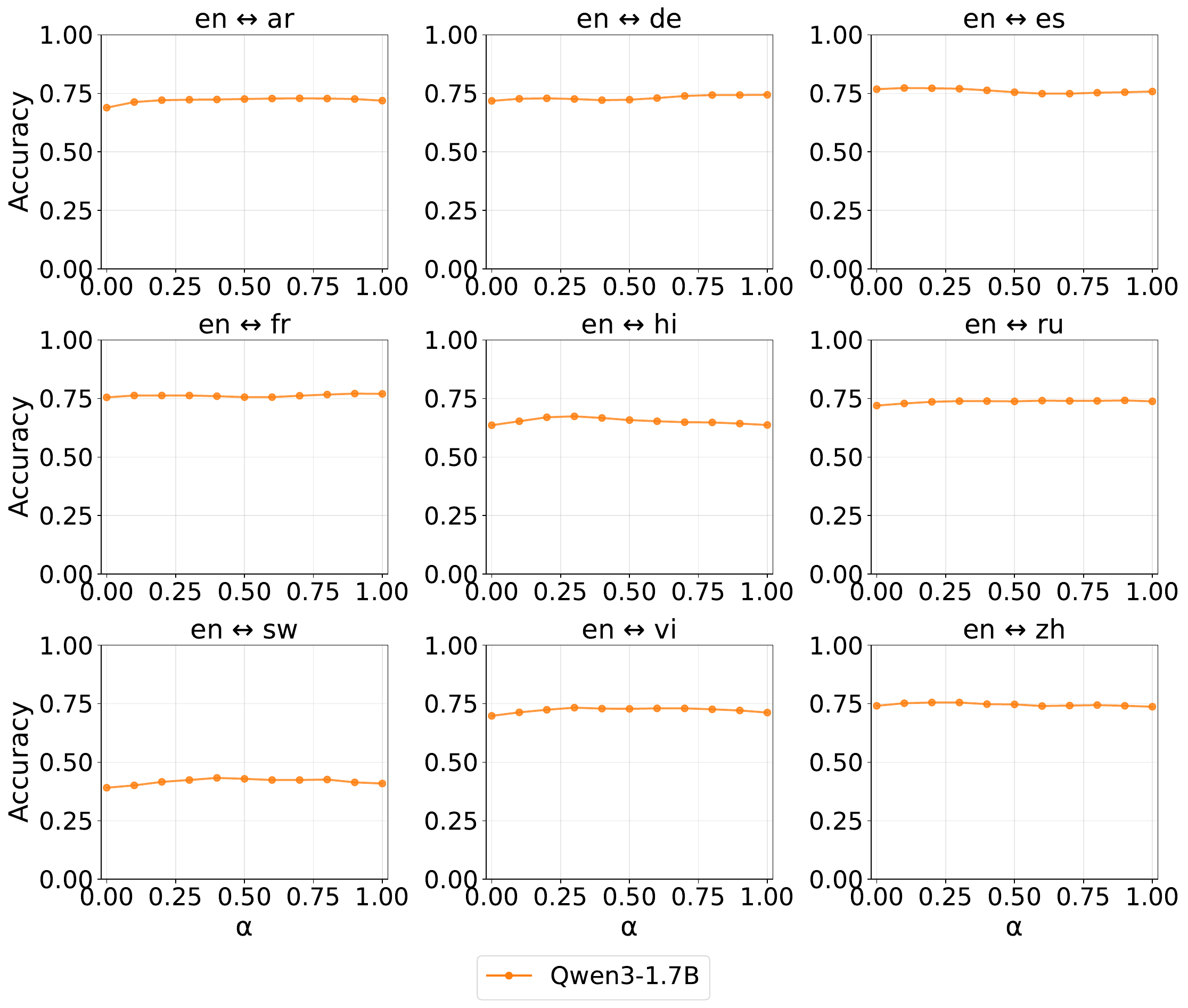}
  \caption{
    Linear interpolation path between English adapter $(\alpha = 0)$ and target-language adapter $(\alpha = 1)$ of Qwen3-1.7B on NLI.
    Each subplot shows accuracy evaluated on XNLI in the respective target language.
  }
  \label{fig:lmc_nli_1b}
\end{figure*}

\begin{figure*}[ht]
  \centering
  \includegraphics[width=\linewidth]{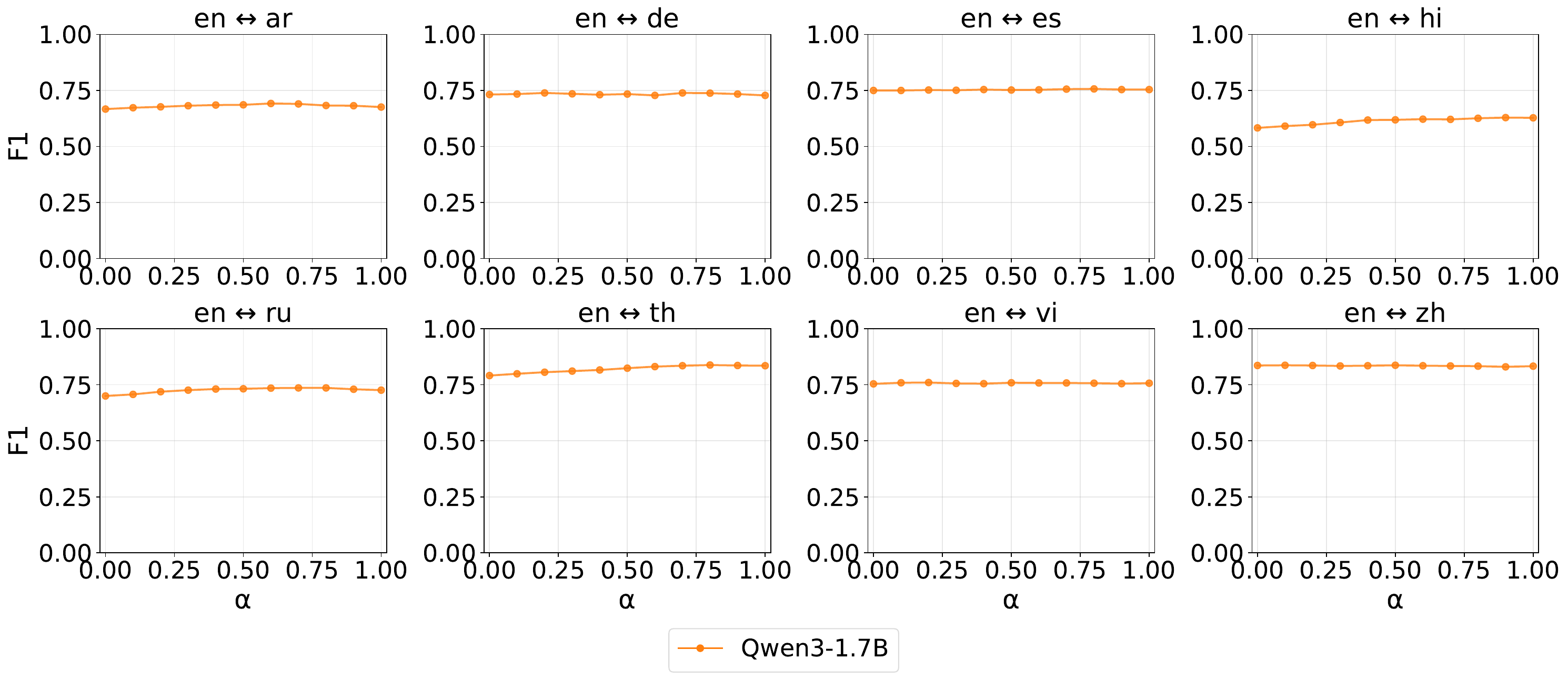}
  \caption{
    Linear interpolation path between English adapter $(\alpha = 0)$ and target-language adapter $(\alpha = 1)$ of Qwen3-1.7B on QA.
    Each subplot shows F1 score evaluated on XQuAD in the respective target language.
  }
  \label{fig:lmc_qa_1b}
\end{figure*}

To demonstrate that our findings generalize to smaller models and to more languages, we conduct additional analysis using Qwen3-1.7B.
For NLI and QA, we use the same languages as in the main text.
For Math, we use the 8 languages that satisfy the language-selection criteria for MGSM as reported in Table~\ref{tab:lang-selection_1b}, covering Bengali (\texttt{bn}), German (\texttt{de}), Spanish (\texttt{es}), French (\texttt{fr}), Japanese (\texttt{ja}), Russian (\texttt{ru}), Swahili (\texttt{sw}), and Thai (\texttt{th}).
To further verify that our findings hold on another Math dataset, we additionally use MSVAMP~\citep{chen2024breaking}, for which we use the 2 languages (\texttt{bn} and \texttt{sw}) following the criterion.
All other settings remain the same as in the main text.

\begin{figure*}[ht]
  \centering
  \includegraphics[width=\linewidth]{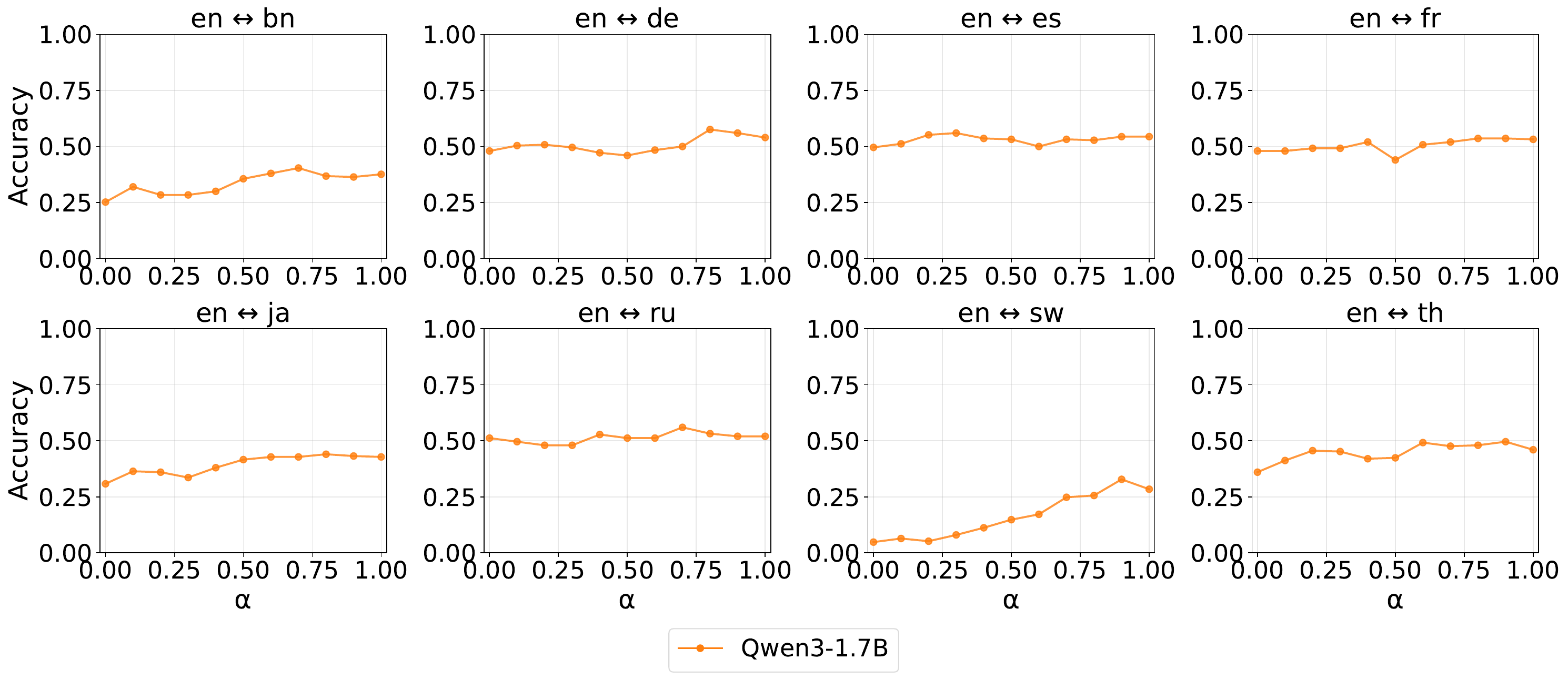}
  \caption{
    Linear interpolation path between English adapter $(\alpha = 0)$ and target-language adapter $(\alpha = 1)$ of Qwen3-1.7B on Math.
    Each subplot shows accuracy evaluated on MGSM in the respective target language.
  }
  \label{fig:lmc_mgsm_1b}
\end{figure*}

\begin{figure*}[ht]
  \centering
  \includegraphics[width=0.5\linewidth]{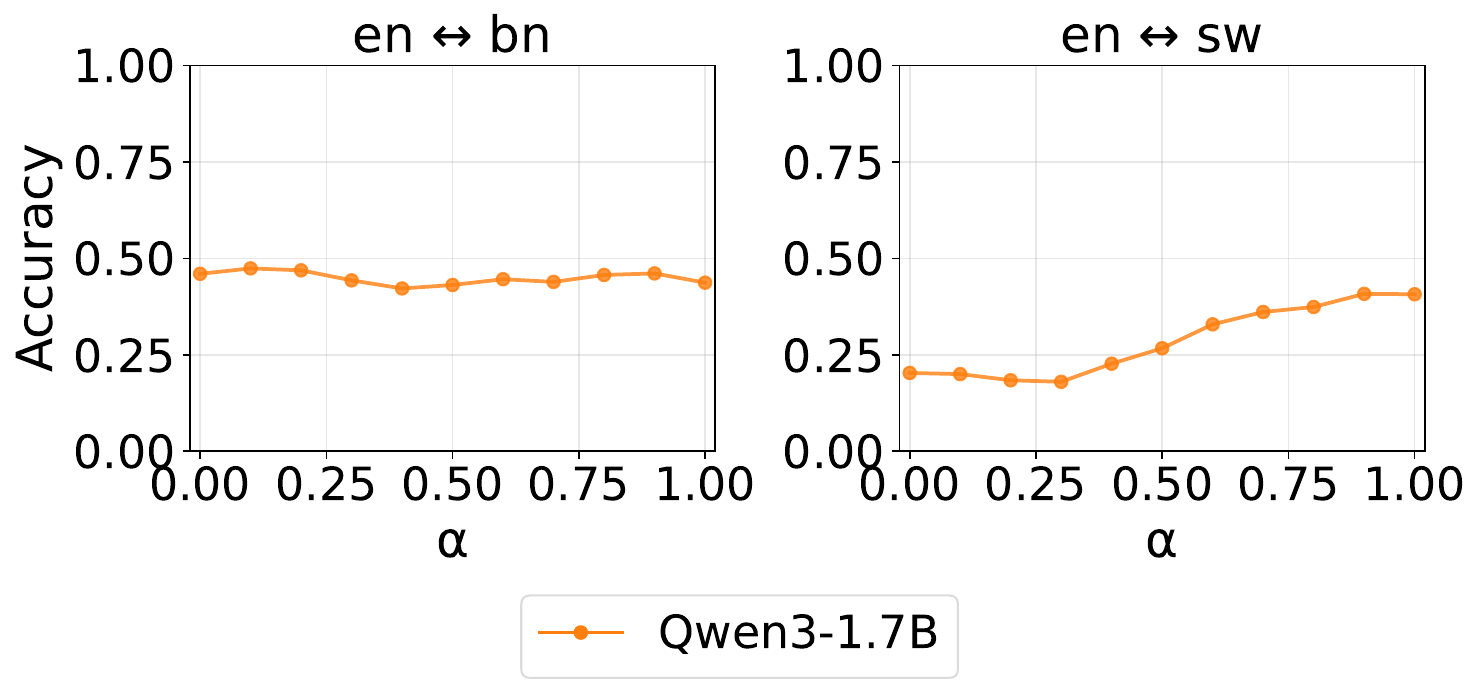}
  \caption{
    Linear interpolation path between English adapter $(\alpha = 0)$ and target-language adapter $(\alpha = 1)$ of Qwen3-1.7B on Math.
    Each subplot shows accuracy evaluated on MSVAMP in the respective target language.
  }
  \label{fig:lmc_msvamp_1b}
\end{figure*}

\paragraph{Cross-lingual linear mode connectivity.}
We observe the same asymmetry in cross-lingual generalization across tasks.
In Figure~\ref{fig:lmc_nli_1b} and Figure~\ref{fig:lmc_qa_1b}, the English adapter and the target-language adapter are linearly connected across all languages on NLI and QA.
In contrast, Figure~\ref{fig:lmc_mgsm_1b} and Figure~\ref{fig:lmc_msvamp_1b} show the opposite behavior, suggesting that this failure of linear connectivity in Math is not confined to a single dataset.
These results indicate that the task-dependent nature of cross-lingual generalization persists even as model size decreases.

\begin{figure*}[ht]
  \centering
  \includegraphics[width=\linewidth]{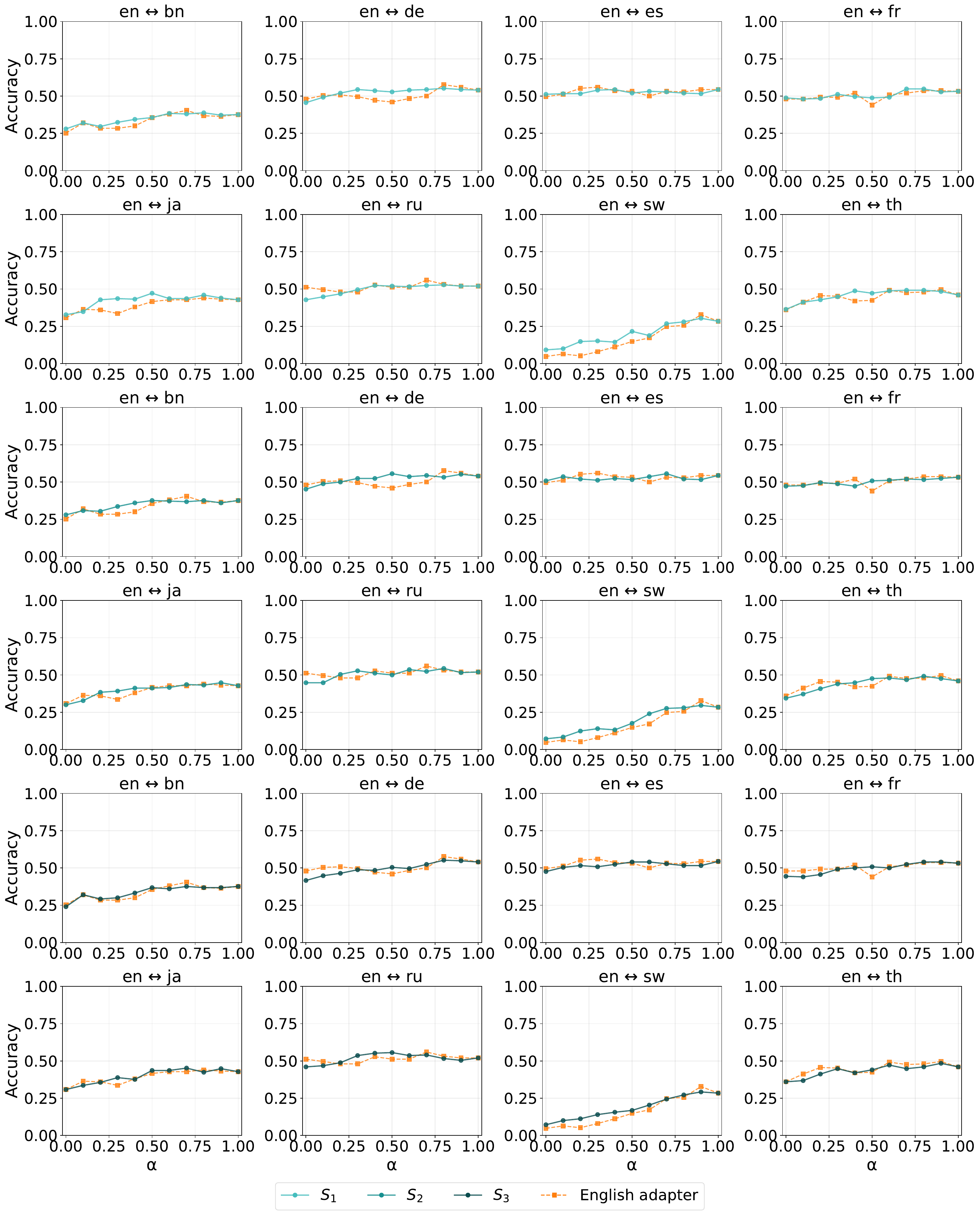}
  \caption{
    Linear interpolation path between ability-adapted English adapter $(\alpha = 0)$ and target-language adapter $(\alpha = 1)$ of Qwen3-1.7B on Math.
    The top (rows 1-2), middle (rows 3-4), and bottom (rows 5-6) row pairs report early ($S_1$), middle ($S_2$), and final ($S_3$) layer adaptation, respectively, covering all eight target languages.
    Each subplot shows accuracy evaluated on MGSM in that language.
  }
  \label{fig:con_lmc_b1}
\end{figure*}

\paragraph{Layer-group localization for cross-lingual generalization.}
We next investigate whether the deficit of cross-lingual generalization on Math is layer-localized in smaller and more diverse languages.
Figure~\ref{fig:con_lmc_b1} shows that, as in Figure~\ref{fig:con_lmc}, adapting only a subset of layers is sufficient to recover LMC in Qwen3-1.7B as well.
Moreover, we observe that adapting early or middle layers is more effective than adapting the final layers.
These results confirm that the conclusion in Section~\ref{sec:R2}, that cross-lingual generalization can be recovered by adapting early and middle layers in this setting.

\paragraph{Factorization without target-language labels.}

\begin{figure*}[ht]
  \centering
  \includegraphics[width=\linewidth]{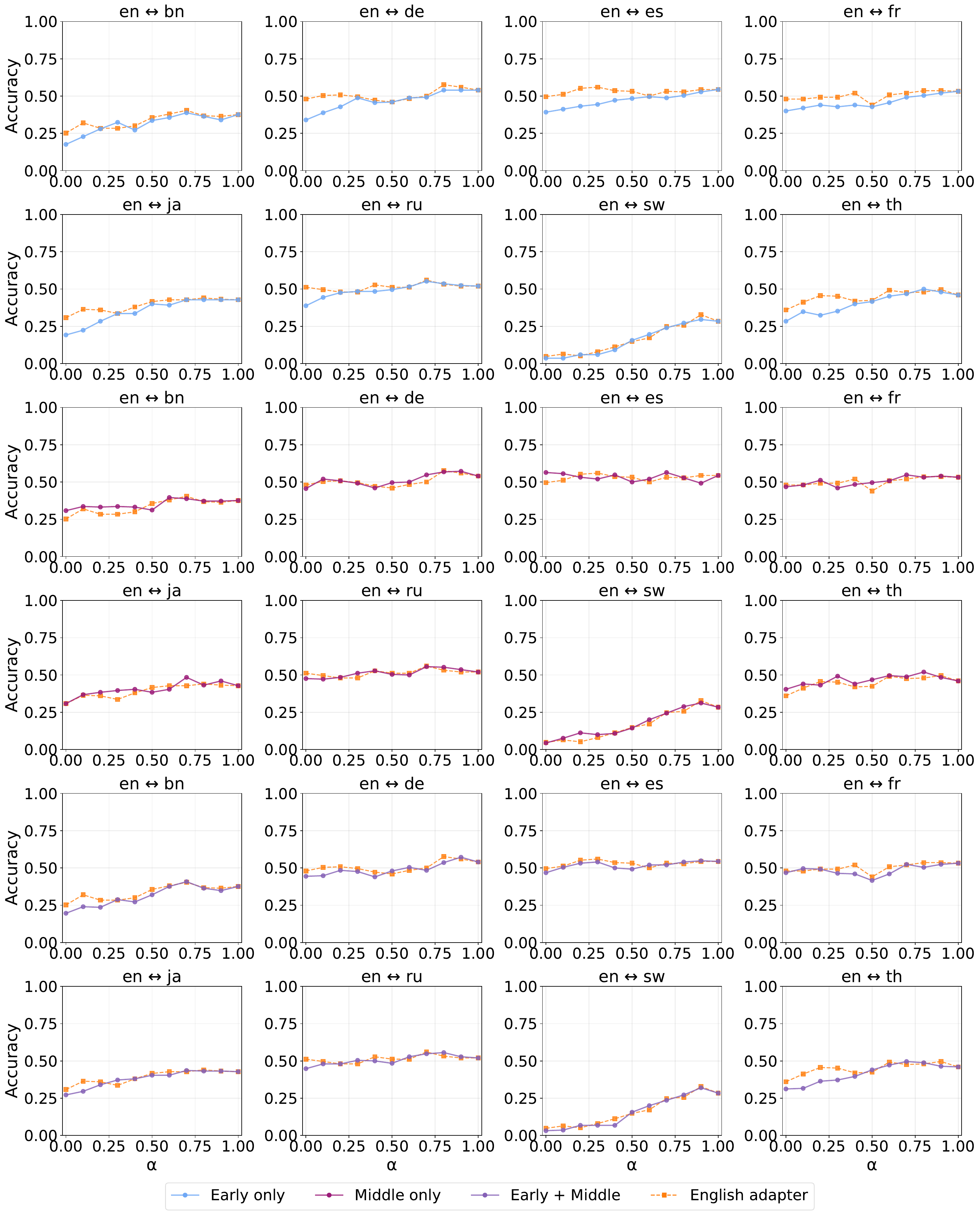}
  \caption{
    Linear interpolation path between routed-adaptation English adapter $\tilde{\boldsymbol{\theta}}_\text{en} (\alpha=0)$ and target-language adapter $\boldsymbol{\theta}_\text{tgt} (\alpha=1)$ on Math.
    The top (rows 1-2), middle (rows 3-4), and bottom (rows 5-6) row pairs report language (Early only), composition (Middle only), and joint (Early and Middle) routing, respectively, covering all eight target languages.
    Each subplot shows accuracy evaluated on MGSM in that language.
  }
  \label{fig:routing_lmc_b1}
\end{figure*}

We further examine whether cross-lingual generalization factorizes in parameter space even for smaller models and a wide range of languages.
Figure~\ref{fig:routing_lmc_b1} and Table~\ref{tab:barrier_1b} show that language routing (Early only) and composition routing (Middle only) both lower the barrier.
Composition routing, in particular, raises accuracy above the naive English adapter.
This mirrors the results in Table~\ref{tab:barrier} and strengthen the conclusion from Section~\ref{sec:R3} that semantic composition is a key factor in cross-lingual generalization.

\end{document}